\documentclass[11pt]{article}

\usepackage[preprint]{acl}
\usepackage{times}
\usepackage{latexsym}
\usepackage[T1]{fontenc}
\usepackage[utf8]{inputenc}
\usepackage{microtype}
\usepackage{inconsolata}
\usepackage{graphicx}
\usepackage{booktabs}
\usepackage{tabularx}
\usepackage{colortbl}
\usepackage{multirow}
\usepackage{multicol}
\usepackage{amssymb}
\usepackage{tabularray}

\definecolor{taxonomyHeader}{HTML}{304B63}
\definecolor{taxonomyPlanning}{HTML}{EDF3FA}
\definecolor{taxonomyExecution}{HTML}{EDF6F2}
\definecolor{taxonomyFeedback}{HTML}{FBF5E9}
\definecolor{taxonomyControl}{HTML}{F2EFF8}

\title{When Jev Helps and Hurts: Delegating High-Level Decisions in LLM Agent Workflows}

\author{
Tongxi Wu\textsuperscript{1} \quad
Kai Chen\textsuperscript{1}\thanks{Kai Chen and Xinfeng Li are co-corresponding authors. Kai Chen: \href{mailto:kaichennjugl@gmail.com}{\texttt{kaichennjugl@gmail.com}}.} \quad
Yukun Wang\textsuperscript{1} \quad
Xinfeng Li\textsuperscript{3}\footnotemark[1] \quad
Yanqi Bao\textsuperscript{2}\thanks{Project leader.} \quad
Yang Gao\textsuperscript{1} \\
\textsuperscript{1}Nanjing University \\
\textsuperscript{2}Horizon Robotics \\
\textsuperscript{3}The Hong Kong Polytechnic University
}

\begin{document}
\maketitle
\begin{abstract}

Repeated planning and verification make LLM agent workflows costly in inference tokens, even for judgments with few possible answers. Jev, a specialized System One model, can handle such judgments to reduce the host's reasoning burden. Existing application-specific studies leave unclear when delegation benefits different hosts and workflows. We organize ten high-level decision functions and evaluate delegation at selected points across six host LLMs on OfficeQA Pro, ALFWorld, and SpreadsheetBench. \textbf{Jev can narrow capability gaps, but delegation gains depend on the host and workflow.} The same decision can help one host and hurt another, and its effect can reverse across workflows. Even when delegation helps, adding more decisions may not improve performance or save host tokens. These gains may also come from workflow changes rather than Jev's judgments. The key is therefore to choose which decisions Jev should make for each host and workflow. Code is available at \url{https://github.com/0xCavaliers/Jev_Workflow}.

\end{abstract}

\section{Introduction}
\label{sec:introduction}

% Declare the comparison table early so it appears at the top of page 2.
\begin{table*}[t]
  \centering
  \small
  \begin{tblr}{
    width=\textwidth,
    colspec={Q[wd=3.15cm] Q[wd=1.3cm] X Q[wd=1.8cm] Q[wd=1.1cm] Q[wd=1.6cm] Q[wd=1.35cm] Q[wd=1.45cm]},
    cells={c,m},
    colsep=2.5pt,
    rowsep=3pt,
    hlines={0.3pt,dashed},
    vline{2-8}={0.3pt,dashed},
    hline{1,2,13,Z}={0.4pt,solid},
  }
    \SetCell[c=2]{c}{\textbf{Approach}} & & \textbf{Decision function} & \textbf{Cross-model\newline evaluation} & \textbf{Effect\newline reversal} & \textbf{Delegation\newline interactions} & \textbf{Workflow\newline controls} & \textbf{Trajectory\newline analysis} \\
    \SetCell[c=2]{c,m}{\shortstack[c]{Edge orchestration\\{\footnotesize\citep{li2026replacing}}}} & & Intent fields & --- & --- & --- & $\checkmark$ & --- \\
    \SetCell[c=2]{c,m}{Jev-Mem~{\footnotesize\citep{jiang2026jev}}} & & Memory / retrieval control & --- & --- & --- & --- & --- \\
    \SetCell[c=2]{c,m}{SeLMRoute~{\footnotesize\citep{perifanis2026selmroute}}} & & Task-demand probes & --- & --- & --- & $\triangle$ & --- \\
    \SetCell[c=2]{c,m}{REFLEX~{\footnotesize\citep{wu2026reflex}}} & & Action gating / fallback & $\checkmark$ & $\triangle$ & --- & $\triangle$ & $\checkmark$ \\
    \SetCell[r=2]{c,m}{\shortstack[c]{JevAgentHarness\\{\footnotesize\citep{zhu2026jevagentharness}}}} & Bench & Tool / action / stop & --- & --- & --- & --- & $\triangle$ \\
    & Harness & Tool / args / reflection & $\checkmark$ & $\triangle$ & $\checkmark$ & $\checkmark$ & $\checkmark$ \\
    \SetCell[r=2]{c,m}{\shortstack[c]{Jev-LiteMAS\\{\footnotesize\citep{wang2026jev}}}} & Router & Expert handoff & $\triangle$ & --- & --- & $\checkmark$ & --- \\
    & Reviewer & Artifact screening & $\triangle$ & --- & $\triangle$ & $\checkmark$ & --- \\
    \SetCell[c=2]{c,m}{Ling et al.~{\footnotesize\citeyearpar{ling2026faster}}} & & Action selection & --- & --- & --- & $\checkmark$ & $\checkmark$ \\
    \SetCell[c=2]{c,m}{Vats~{\footnotesize\citeyearpar{vats2026good}}} & & Stop / verify / recover & --- & --- & --- & $\checkmark$ & $\triangle$ \\
    \SetCell[c=2]{c,m}{\parbox[c]{4cm}{\centering AgenticJevNav\\{\footnotesize\citep{zhang2026agenticjevnav}}\par}} & & Skill / failure / recovery & --- & --- & --- & $\checkmark$ & $\checkmark$ \\
    \SetCell[c=2]{c}{\textbf{Ours}} & & \textbf{High-level functions} & $\checkmark$ & $\checkmark$ & $\checkmark$ & $\checkmark$ & $\checkmark$ \\
  \end{tblr}
  \par\vspace{3pt}
  \begin{minipage}{\textwidth}
    \footnotesize
    $\checkmark$: explicit analysis; $\triangle$: partial coverage; ---: not an analysis focus. Numbers count hosts evaluated for the same function.
  \end{minipage}
  \caption{Comparison of decision functions and empirical analyses in Jev-based systems.}
  \label{tab:related-work-comparison}
\end{table*}

Agents built on large language (LLMs) automate tasks such as web navigation, document question answering, and spreadsheet editing~\citep{zhou2024webarena,opsahl2026officeqa,ma2024spreadsheetbench}. These workflows interleave reasoning and tool use~\citep{yao2022react}, requiring repeated decisions about information gathering, plan revision, and result verification. Invoking the primary LLM for each judgment adds token overhead~\citep{kim2023llm,kim2026cost}, even when only a few answers are possible. Jev, a specialized System One model, can handle these bounded judgments through structured choices and probabilities while the host LLM handles generation~\citep{almeida2026jev,deusser2026evaluating}.

Existing studies apply Jev to classification, routing, and scoring~\citep{deusser2026evaluating}, and integrate it into agent control through confidence-based delegation~\citep{zhu2026jevagentharness,wu2026reflex}. \textbf{Table}~\ref{tab:related-work-comparison} summarizes the empirical analyses reported in these studies. Although delegation offers cheaper local judgments, whether these savings translate into more efficient task execution remains an empirical question~\citep{ling2026faster,vats2026good}. We focus on \textbf{high-level decision functions}: bounded judgments that govern how a workflow proceeds, such as whether to gather more information, revise a plan, or accept a result. The key question is \textbf{whether delegating these judgments reduces host LLM token use while maintaining or improving task performance}.

To answer this question, we compare decision delegation across \textbf{six host LLMs and three workflows}. Drawing on existing Jev applications, we organize ten candidate decision functions across planning, execution, feedback, and control (Table~\ref{tab:replacement-points}). We evaluate selected functions individually and in combination on OfficeQA Pro~\citep{opsahl2026officeqa}, ALFWorld~\citep{shridhar2020alfworld}, and SpreadsheetBench~\citep{ma2024spreadsheetbench}. We include workflow baselines without Jev and controls where the host LLM makes the same decisions. These comparisons help identify whether improvements come from Jev's judgments or from changes to the workflow. We assess task performance, host LLM token use, and execution behavior.

Our results show that \textbf{delegation can improve task performance without consistently reducing host LLM token use}. It can compensate for a host's decision weaknesses, yet the same decision function can help one host and hurt another, and its benefits can change across workflows. Individual gains also do not reliably predict combined performance, because delegated decisions can support or disrupt one another during execution. Interpreting these gains further requires workflow controls: added processing steps can account for improvements otherwise attributed to Jev. The central challenge is therefore to determine \textbf{which decisions a host should retain and which it should delegate within a given workflow}.

Our contributions can be summarized as follows.
\begin{itemize}
  \item \textbf{We organize high-level decision functions into a common taxonomy.} Drawing on existing Jev applications, we identify ten candidate functions across planning, execution, feedback, and control.
  \item \textbf{We compare decision delegation across six host LLMs and three workflows.} Individual and combined delegation settings are evaluated against workflow baselines and host-LLM decision controls.
  \item \textbf{We provide empirical guidance for allocating decision authority.} We show that delegation benefits depend on the host and workflow, and that individual performance gains do not reliably predict combined performance or host-token savings.
\end{itemize}

\section{Related Work}
\label{sec:related_work}

% Declare the taxonomy table early so it appears above Section 3 on page 3.
\begin{table*}[t]
  \centering
  \small
  \renewcommand{\arraystretch}{1.18}
  \setlength{\tabcolsep}{4pt}
  \captionsetup{justification=centering}
  \renewcommand{\tabularxcolumn}[1]{m{#1}}
  \begin{tabularx}{\textwidth}{c c >{\centering\arraybackslash}m{3.7cm} >{\raggedright\arraybackslash}X c}
    \arrayrulecolor{taxonomyHeader}
    \toprule
    \rowcolor{taxonomyHeader}
    \textcolor{white}{\textbf{Stage}} & \textcolor{white}{\textbf{ID}} & \multicolumn{1}{c}{\cellcolor{taxonomyHeader}\textcolor{white}{\textbf{Replacement point}}} & \multicolumn{1}{c}{\cellcolor{taxonomyHeader}\textcolor{white}{\textbf{Description}}} & \textcolor{white}{\textbf{Jev interface}} \\
    \midrule
    \rowcolor{taxonomyPlanning}
    & 1 & Strategy Selection & Choose a task-solving strategy, such as retrieval, computation, direct reasoning, or interaction. & Choice \\
    \rowcolor{taxonomyPlanning}
    & 2 & Plan / Replan Decision\par{\footnotesize\citep{zhang2026agenticjevnav}} & Decide whether to adjust or replace the current plan. & Noul \\
    \rowcolor{taxonomyPlanning}
    \multirow{-3}{*}[1ex]{\textbf{Planning}} & 3 & Task Requirement Interpretation\par{\footnotesize\citep{li2026replacing,perifanis2026selmroute}} & Identify key constraints and the expected output format to select subsequent processing rules. & Choice \\
    \midrule
    \rowcolor{taxonomyExecution}
    & 4 & Tool-use Decision\par{\footnotesize\citep{wu2026reflex}} & Decide whether the current step requires an external tool. & Noul \\
    \rowcolor{taxonomyExecution}
    & 5 & Tool Selection\par{\footnotesize\citep{zhu2026jevagentharness}} & Select a suitable tool from the predefined tool set. & Choice \\
    \rowcolor{taxonomyExecution}
    \multirow{-2.5}{*}[1.5ex]{\textbf{Execution}} & 6 & Information Acquisition Strategy\par{\footnotesize\citep{jiang2026jev}} & Choose what type of information to acquire or inspect next, rather than generate search terms or action parameters. & Choice \\
    \midrule
    \rowcolor{taxonomyFeedback}
    & 7 & Information Sufficiency\par{\footnotesize\citep{jiang2026jev}} & Assess whether current evidence supports the next operation or more information is needed. & Noul / Score \\
    \rowcolor{taxonomyFeedback}
    & 8 & Result Verification\par{\footnotesize\citep{wang2026jev}} & Assess whether the execution result meets task requirements and the intended goal. & Noul / Score \\
    \rowcolor{taxonomyFeedback}
    \multirow{-4}{*}[0.5ex]{\textbf{Feedback}} & 9 & Failure Recovery\par{\footnotesize\citep{zhang2026agenticjevnav}} & On failure, choose to retry, change strategy, or replan. & Choice \\
    \midrule
    \rowcolor{taxonomyControl}
    \textbf{Control} & 10 & Termination / Escalation\par{\footnotesize\citep{wu2026reflex}} & Decide whether to continue, terminate, or escalate the task. & Choice \\
    \bottomrule
  \end{tabularx}
  \caption{Ten candidate high-level replacement points across four workflow stages. Citations identify related decision functions in prior Jev applications. The final column lists possible Jev interface formulations.}
  \label{tab:replacement-points}
\end{table*}

\textbf{Reasoning and Decision-Making in Agent Workflow.} LLM agent workflows combine reasoning, action, and feedback to solve multi-step tasks. ReAct interleaves reasoning with actions and observations~\citep{yao2022react}, while LATS integrates search, planning, and environmental feedback~\citep{zhou2023language}. Reflexion uses verbal feedback to guide subsequent attempts~\citep{shinn2023reflexion}, and Self-Refine improves outputs through iterative self-feedback and revision~\citep{madaan2023self}. These iterative reasoning and feedback loops also incur token overhead~\citep{kim2026cost}, motivating more selective use of LLM inference within agent workflows.

\textbf{Specialized Decision Models.} Jev supports classification, routing, and other bounded decisions through structured choices, scores, and probabilities~\citep{almeida2026jev,deusser2026evaluating}. Applications include text annotation, legal document analysis, evaluation across languages, and confidence-based judge routing, among other uses~\citep{ibrahim2026evaluating,zhang2026same,liu2026jev,li2026jev}. Jev-Mem extends this use to agentic memory, assigning Jev control over memory organization, retrieval routing, budget allocation, and stopping~\citep{jiang2026jev}. These studies establish Jev's utility within particular tasks or subsystems, but their application-specific results leave open which high-level decision functions remain useful across workflows and host LLMs. We examine these differences together with interactions between delegated functions.

\textbf{Decision Delegation in Agent Workflows.} JevAgentHarness studies decision points, integration modes, and calibrated delegation~\citep{zhu2026jevagentharness}. REFLEX examines confidence-gated control~\citep{wu2026reflex}, while Jev-LiteMAS combines expert handoffs with selective review~\citep{wang2026jev}. Other studies examine efficiency transfer and controller interventions~\citep{ling2026faster,vats2026good}, or decision models as actors, judges, and teachers~\citep{liu2026actor}. Table~\ref{tab:related-work-comparison} summarizes the scope of these analyses. We examine how high-level decision authority should be divided between Jev and the host LLM. Across six hosts and three workflows, we compare individual and combined delegation settings, using workflow controls and execution traces to explain when delegation helps or disrupts task execution.

\section{High-Level Decision Delegation}
\label{sec:framework}

To compare delegation across workflows, we first distinguish \emph{the functions of the decisions being delegated}. Table~\ref{tab:replacement-points} defines ten candidate high-level replacement points across planning, execution, feedback, and control. The taxonomy groups judgments by their purpose: information acquisition (\#6) concerns what to inspect next, whereas tool selection (\#5) concerns which tool to use. Likewise, information sufficiency (\#7) assesses readiness to proceed, while result verification (\#8) assesses the output already produced. These distinctions provide a common vocabulary for comparing decisions across tasks and host models.

We examine an agent's procedures and rules to identify decisions that can be expressed as bounded choices or assessments. We map each decision to the taxonomy according to its purpose and specify \textbf{when Jev is consulted, what it receives, and how its judgment changes execution}. A trigger may be an agent request, a workflow stage transition, or an observed failure. The controller supplies the relevant task context, observations or execution evidence, and candidate options or assessment criteria. It then translates Jev's response into a selected action, routing instruction, or acceptance decision. The concrete encoding depends on the workflow: a decision may select an admissible environment action directly or guide further LLM generation. The host retains responsibility for the portions of the task outside the delegated point.

Once a decision is assigned to Jev, we can examine how that delegation affects the agent's performance and behavior. We also delegate multiple decisions within the same workflow to examine whether they work well together. Section~\ref{sec:setup} describes which decisions we delegate in each task and how they are implemented.

\section{Experimental Setup}
\label{sec:setup}

\begin{figure*}[t]
  \centering
  % The compile copy removes embedded draw.io XML; regenerate after re-exporting.
  % Run: python3 scripts/prepare_method_figure.py
  % Crop the A4 export to the method diagram, retaining a 1 bp margin.
  \includegraphics[width=\textwidth,trim=18.62bp 433.99bp 151.07bp 255.79bp,clip]{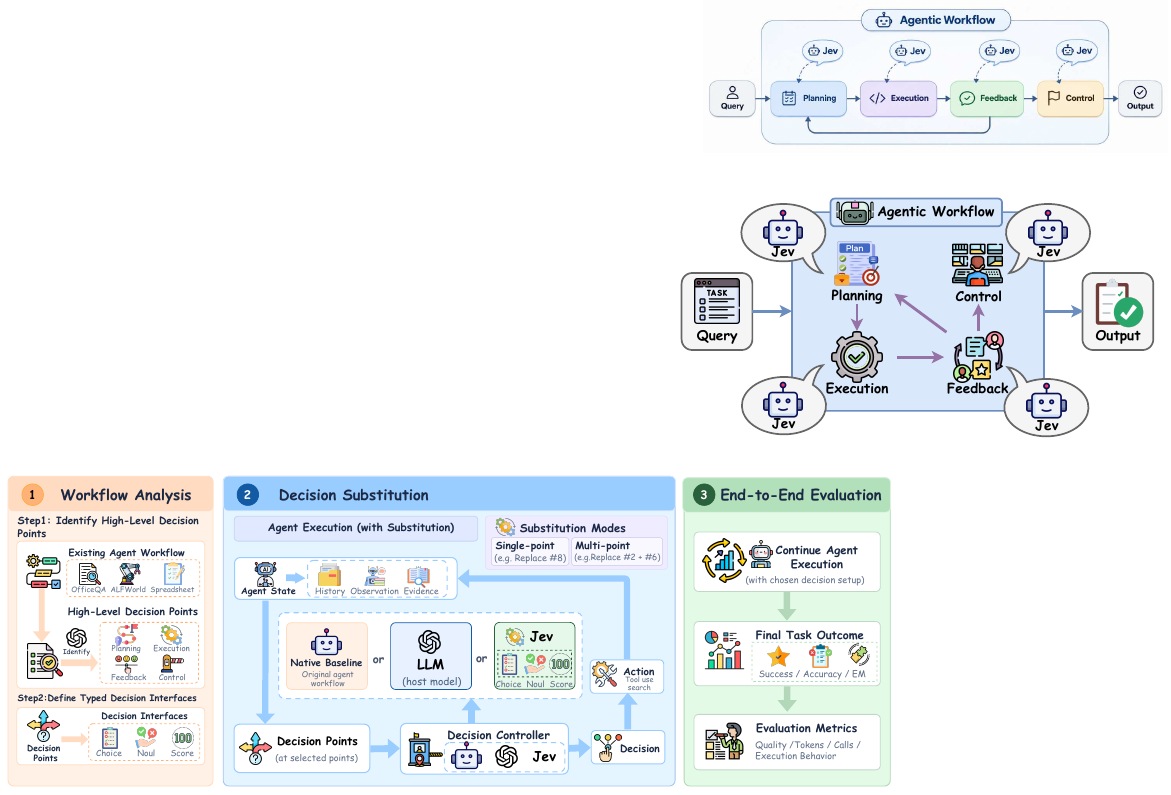}
  \caption{Overview of our high-level decision delegation and evaluation protocol.}
  \label{fig:jev_method}
\end{figure*}

\textbf{Study overview.} Figure~\ref{fig:jev_method} illustrates our study in three stages. We first use GPT-6.1 Sol~\citep{openai2026gpt61sol} to inspect the existing \emph{skills} in SkillOpt~\citep{yang2026skillopt}, which describe task-solving procedures and rules, and identify candidate replacement points under our taxonomy. We then implement selected points in the workflows and delegate their judgments to Jev, individually or in combination. Finally, we compare task outcomes, resource use, and execution behavior against native workflows and available host-LLM decision controls.

\textbf{Task benchmarks.} We select three benchmarks covering document question answering, text-based embodied tasks, and spreadsheet manipulation:
\begin{itemize}
  \begin{samepage}
  \item \textbf{OfficeQA Pro}~\citep{opsahl2026officeqa}: 133 document questions involving evidence lookup and numerical reasoning, with oracle source pages and candidate files provided.\par
  \end{samepage}
  \item \textbf{ALFWorld}~\citep{shridhar2020alfworld}: 134 unseen tasks across six types, requiring navigation and object manipulation in a text-based embodied environment.
  \item \textbf{SpreadsheetBench}~\citep{ma2024spreadsheetbench}: 280 workbook editing tasks specified in natural language, solved with Python and evaluated on task-specific test workbooks.
\end{itemize}

\begin{table}[t]
  \centering
  \small
  \renewcommand{\arraystretch}{1.12}
  \setlength{\tabcolsep}{4pt}
  \arrayrulecolor{black}
  \captionsetup{justification=centering}
  \renewcommand{\tabularxcolumn}[1]{m{#1}}
  \begin{tabularx}{\columnwidth}{@{}>{\centering\arraybackslash}m{2.8cm} >{\raggedright\arraybackslash}X@{}}
    \toprule
    \textbf{Dataset} & \multicolumn{1}{c@{}}{\textbf{Jev decision}} \\
    \midrule
    OfficeQA Pro & \#3 Task interpretation\newline
      \#7 Evidence sufficiency\newline
      \#8 Result verification \\
    \midrule
    ALFWorld & \#2 Actions while holding an object\newline
      \#3 Task classification\newline
      \#6 Search action selection\newline
      \#7 Object pickup or abstention\newline
      \#9 Recovery after failed actions \\
    \midrule
    SpreadsheetBench & \#6 Workbook inspection\newline
      \#7 Readiness to write code\newline
      \#8 Output verification \\
    \bottomrule
  \end{tabularx}
  \caption{Single-point delegation settings.}
  \label{tab:task-replacements}
\end{table}

\textbf{Model selection.} To examine how delegation depends on the host model, we evaluate GPT-5.4 mini~\citep{openai2026gpt54mini}, GPT-6.1 Sol~\citep{openai2026gpt61sol}, GPT-6 Luna~\citep{openai2026gpt6solluna}, DeepSeek-v4.1-flash~\citep{xu2026deepseek}, Claude Haiku 5.5~\citep{anthropic2026haiku55} and Qwen3.8-Flash~\citep{qiu2026design}, each with its provider's default reasoning effort.

\textbf{Single-point evaluation.} For each setting in Table~\ref{tab:task-replacements}, we delegate one decision function to Jev~\citep{almeida2026jev} through the \texttt{jev-latest} backend and leave the remaining decisions to the host. For each workflow, the baseline and delegation settings share the same task set and execution budget: 24 model rounds for OfficeQA Pro, 50 environment steps for ALFWorld, and 30 interaction turns for SpreadsheetBench.

\textbf{Exploratory combinations.} After examining the single-point results, we evaluate selected combinations to investigate whether delegation benefits persist when multiple decision functions are delegated together. This exploratory evaluation covers a subset of possible combinations.

\textbf{Baselines and controls.} We compare each setting with the available workflow baselines and decision controls. In OfficeQA Pro, the controls include the evidence-ledger workflow without Jev, code recomputation alone where available, and recomputation with host-LLM verification. For SpreadsheetBench, host-LLM verification uses the same evidence and revision flow as Jev. In ALFWorld, Jev sees the full episode history, while the native host sees only the two most recent steps. To assess the effect of this additional context, we also evaluate a host-only baseline with full history.

\textbf{Evaluation metrics.} We report normalized EM and token-level F1 for OfficeQA Pro, environment-goal success rate for ALFWorld, and task success rate for SpreadsheetBench, which requires passing the output-cell checks on all test workbooks. We measure host LLM tokens and model calls per task, recording Jev tokens and calls separately. Workflow-specific statistics, including repeated actions and tool calls, describe how delegation changes execution.

\section{Empirical Results}
\label{sec:results}

We present five findings on when high-level decision delegation improves task performance, how it affects host-token use, and what these effects imply for allocating decision authority.

\subsection{Capability Amplification through Decision Delegation}
\label{sec:amplification}

\begin{table*}[t]
  \centering
  \small
  \setlength{\tabcolsep}{2pt}
  \renewcommand{\arraystretch}{1.12}
  \arrayrulecolor{black}
  \begin{tabular*}{\textwidth}{@{\hspace{6pt}\extracolsep{\fill}}lcccccccccccccc@{}}
    \toprule
    \multicolumn{1}{c}{\multirow{2}{*}{Host model}} & \multicolumn{3}{c}{Success (\%)} & \multicolumn{3}{c}{Env. steps} & \multicolumn{3}{c}{Ineffective actions} & \multicolumn{3}{c}{LLM tokens (k)} & \multicolumn{2}{c}{Jev tokens (k)} \\
    \cmidrule(lr){2-4}\cmidrule(lr){5-7}\cmidrule(lr){8-10}\cmidrule(lr){11-13}\cmidrule(lr){14-15}
    & LLM & \#6 & \#6+\#2 & LLM & \#6 & \#6+\#2 & LLM & \#6 & \#6+\#2 & LLM & \#6 & \#6+\#2 & \#6 & \#6+\#2 \\
    \midrule
    GPT-5.4 mini & 79.85 & 89.55 & \textbf{97.01} & 22.09 & 18.11 & \textbf{13.04} & 2.04 & 0.34 & \textbf{0.03} & 88.65 & 32.16 & \textbf{12.53} & 34.12 & 36.07 \\
    Qwen3.8-Flash & \textbf{97.76} & \textbf{97.76} & 97.01 & \textbf{10.59} & 12.31 & 12.77 & 0.07 & 0.01 & \textbf{0.00} & 45.83 & 28.70 & \textbf{16.36} & 25.16 & 37.04 \\
    DeepSeek-v4.1-flash & \textbf{100.00} & 99.25 & \textbf{100.00} & \textbf{9.97} & 11.24 & 11.46 & 0.07 & 0.01 & \textbf{0.00} & 38.09 & 21.52 & \textbf{13.58} & 22.47 & 33.53 \\
    GPT-6 Luna & \textbf{98.51} & \textbf{98.51} & \textbf{98.51} & 11.58 & \textbf{11.44} & 11.96 & 0.06 & \textbf{0.00} & \textbf{0.00} & 43.75 & 20.75 & \textbf{12.20} & 22.76 & 34.30 \\
    GPT-6.1 Sol & \textbf{100.00} & \textbf{100.00} & 99.25 & \textbf{9.21} & 11.13 & 11.61 & 0.02 & \textbf{0.00} & \textbf{0.00} & 64.77 & 105.16 & \textbf{46.65} & 22.63 & 34.04 \\
    Claude Haiku 5.5 & \textbf{97.01} & 94.78 & \textbf{97.01} & 13.87 & 14.12 & \textbf{12.95} & 0.23 & 0.25 & \textbf{0.01} & 62.20 & 31.43 & \textbf{16.67} & 25.61 & 37.71 \\
    \bottomrule
  \end{tabular*}
  \caption{ALFWorld results on 134 tasks. LLM denotes the full-history host-only control; \#6 delegates search, and \#6+\#2 adds held-object progression. Steps, ineffective actions (no environment effect), and tokens are per-task averages. Bold marks the highest success and lowest steps, ineffective actions, and LLM token use per host.}
  \label{tab:alfworld-amplification}
\end{table*}

\begin{figure*}[t]
  \noindent\includegraphics[width=0.94\textwidth]{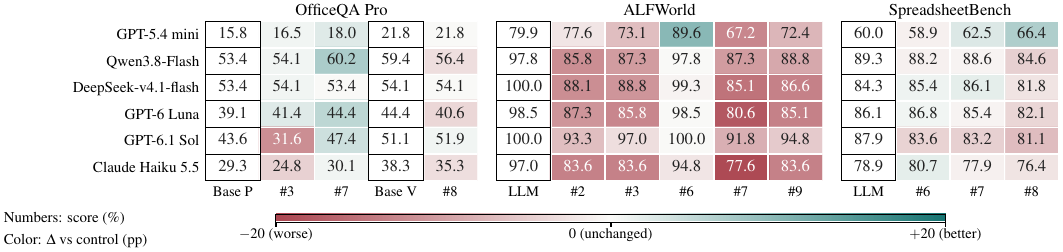}
  \caption{Single-point delegation. Numbers: EM or success (\%); colors: change (pp) from white baseline columns. OfficeQA uses protocol control P for \#3/\#7 and LLM verification V for \#8; ALFWorld uses full-history LLM, and SpreadsheetBench uses native ReAct. Point IDs follow Table~\ref{tab:replacement-points}.}
  \label{fig:delegation-effects}
\end{figure*}

\textbf{Jev can narrow capability gaps by correcting decisions that stall task progress.} On ALFWorld, GPT-5.4 mini achieves 79.85\% success with full-history LLM decisions. Delegating search (\#6) to Jev raises success to 89.55\%, and the selected combination of search and held-object progression (\#6+\#2) reaches 97.01\% (Table~\ref{tab:alfworld-amplification}). This 17.16 percentage-point gain brings mini close to GPT-6.1 Sol's 100\% success under the full-history control. The improvement also accompanies a reduction in ineffective actions from 2.04 to 0.03 per task, indicating more successful execution with fewer actions that leave the environment unchanged.

The trajectories show why finding an object is only part of solving the task. In several failed full-history episodes, the host has already found and picked up the required object, yet repeatedly puts it down or tries to clean it before reaching a sink. In the corresponding Jev runs, the agent carries the object to the sink, cleans it, and returns it to the destination, turning an unfinished sequence into a completed task. GPT-6.1 Sol and DeepSeek already follow these sequences with few ineffective actions (Table~\ref{tab:alfworld-amplification}), so delegating the same decisions offers fewer opportunities to recover stalled progress. These traces suggest that \emph{capability amplification can come from correcting decisions that otherwise prevent the agent from advancing to its next subgoal.}

\subsection{Delegation Effects Reverse across Hosts and Workflows}
\label{sec:inversion}

\textbf{The same delegated decision can help one host and hurt another, with effects that also vary across workflows.} On SpreadsheetBench, Jev verification (\#8) raises GPT-5.4 mini's success from 60.00\% to 66.43\%, but lowers GPT-6.1 Sol's from 87.86\% to 81.07\% (Figure~\ref{fig:delegation-effects}). This reversal extends beyond the pair: verification lowers success for all five hosts other than mini. For the same host, the effect can also reverse across workflows: verification improves GPT-6.1 Sol's OfficeQA Pro EM by 0.75 percentage points relative to host-LLM verification, but lowers its SpreadsheetBench success by 6.79 points relative to native ReAct. The choice of decision function also matters: on SpreadsheetBench, DeepSeek improves with information sufficiency (\#7), yet declines with verification (\#8).

The traces show how a request to revise can either rescue an unfinished answer or undo a valid choice. In one mini task, Jev refuses to accept a cell containing an unevaluated formula; the host continues and eventually writes the required concrete value. In a Sol task, Jev rejects an output containing intentionally blank cells. The resulting feedback prompts the host to fill them with ``Keep'', introducing values where the task requires blanks and causing the final output to fail. These examples suggest that \emph{delegation helps when it corrects the host's omissions, but can hurt when it induces revisions to decisions that were already correct.}

\begin{figure*}[t]
  \noindent\makebox[\textwidth][l]{\hspace*{-0.35in}\includegraphics[width=\textwidth]{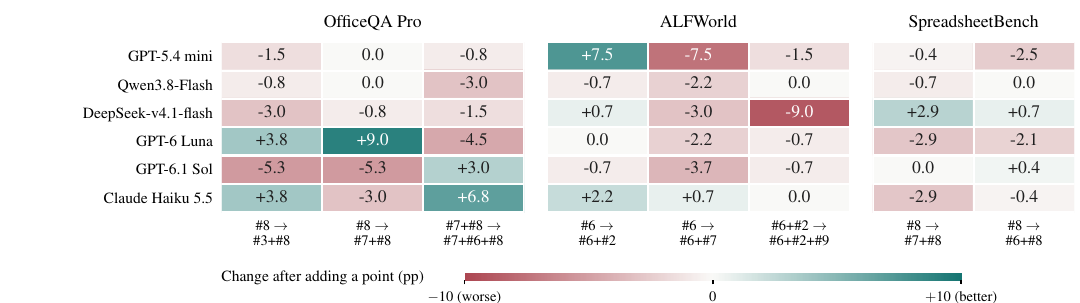}}
  \caption{Incremental effects of extending delegation. Numbers and colors show the change in EM or success (pp) from the configuration above each arrow to the configuration below it.}
  \label{fig:delegation-interactions}
\end{figure*}

\begin{figure*}[t]
  \noindent\makebox[\textwidth][l]{\hspace*{-0.35in}\includegraphics[width=\textwidth]{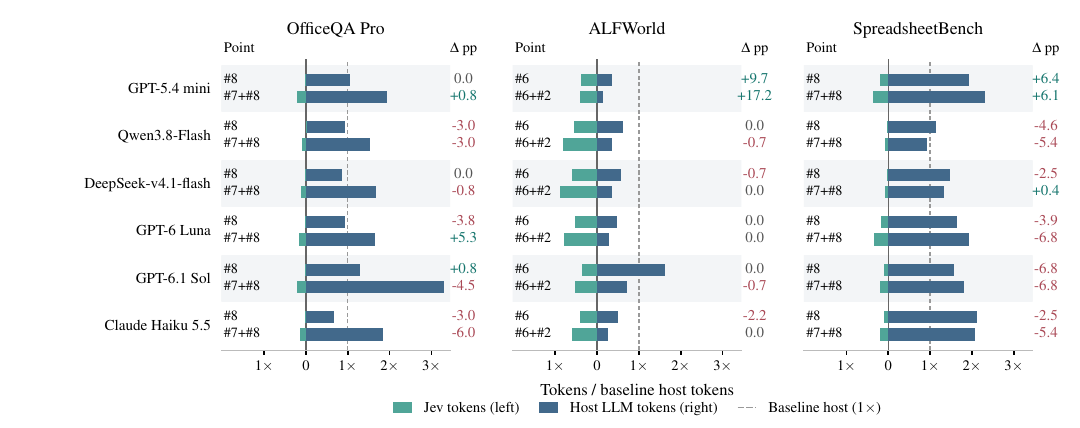}}
  \caption{Token allocation and task performance. Each row labels its delegated decision points. Bars show Jev tokens to the left and host tokens to the right, normalized by baseline host tokens; the adjacent column reports the change in EM or success (pp). Baselines are LLM verification for OfficeQA, full-history LLM for ALFWorld, and native ReAct for SpreadsheetBench.}
  \label{fig:delegation-resources}
\end{figure*}

\subsection{Disentangling Delegation Gains from Workflow Effects}
\label{sec:workflow-effects}

\textbf{Gains after introducing Jev may come from workflow changes rather than Jev's judgments.} In the matched OfficeQA Pro comparison, GPT-5.4 mini answers 20 questions correctly with the protocol-only workflow. Giving the workflow a code recomputation step raises this count to 28, and adding either host-LLM or Jev verification reaches 29 (Table~\ref{tab:officeqa-attribution}). Most of the EM increase appears when the workflow gains a way to recompute its answer, before either model makes a verification judgment.

A similar distinction emerges when delegation gives a decision maker more context. On ALFWorld, extending mini's host history from two steps to the full episode increases completed tasks from 96 to 107 without Jev. The \#6+\#2 configuration reaches 130, so its advantage persists against the full-history control, but comparison with the short-history baseline would also credit delegation for the benefit of additional context. Because the delegation configuration also reallocates task rules, the remaining gap describes the configuration as a whole. Together, these comparisons show that \emph{workflow improvements must be accounted for before attributing an observed gain to Jev's judgments.}

\begin{table}[t]
  \centering
  \small
  \setlength{\tabcolsep}{5pt}
  \begin{tabular}{lrr}
    \toprule
    Configuration & Correct & EM (\%) \\
    \midrule
    Protocol only & 20 & 15.04 \\
    Code recomputation & 28 & 21.05 \\
    Code + LLM verification & 29 & 21.80 \\
    Code + Jev verification & 29 & 21.80 \\
    \bottomrule
  \end{tabular}
  \caption{Matched OfficeQA Pro attribution results for GPT-5.4 mini on 133 questions.}
  \label{tab:officeqa-attribution}
\end{table}

\subsection{Individual Gains Do Not Reliably Predict Combined Performance}
\label{sec:interactions}

\textbf{Individual gains do not reliably predict how delegated decisions perform together.} On ALFWorld, GPT-5.4 mini reaches only 77.61\% success with held-object progression (\#2) alone, below its 79.85\% full-history baseline. Yet combining \#2 with search (\#6) raises success from 89.55\% to 97.01\%; combining search with pickup judgments (\#7) instead lowers it to 82.09\% (Figures~\ref{fig:delegation-effects} and~\ref{fig:delegation-interactions}). On OfficeQA Pro, GPT-6 Luna's \#7+\#8 exceeds either point alone, but adding information acquisition (\#6) reduces EM from 49.62\% to 45.11\%. Thus, neither an individual point's score nor a successful smaller combination determines the performance of a larger one.

The alarmclock task illustrates why decisions must work together through execution. With search alone, the agent reaches the desk, picks up the alarmclock, and uses the desk lamp to finish. With \#7 added, it reaches the same desk, but later pickup checks repeatedly decline to take the visible object. Examination and lamp-use actions continue until the step limit. By contrast, \#2 can help turn a successful search into completed transformation and delivery steps, as illustrated in Section~\ref{sec:amplification}. The added decision can therefore either carry forward or block progress established by another. These cases suggest that \emph{combined performance depends on how delegated decisions work together, not simply on how well each performs alone.}

\subsection{Performance Gains Do Not Imply Host-Token Savings}
\label{sec:resource-use}

\textbf{Higher task success does not guarantee lower host-token use.} On ALFWorld, GPT-5.4 mini's \#6+\#2 improves success from 79.85\% to 97.01\% while reducing host tokens from 88.65k to 12.53k per task relative to the full-history control (Table~\ref{tab:alfworld-amplification}). On SpreadsheetBench, mini's verification (\#8) improves success from 60.00\% to 66.43\%, but increases host tokens from 26.4k to 51.0k. The configurations also use 36.07k and 5.3k Jev tokens per task, respectively (Figure~\ref{fig:delegation-resources}). Their shared performance improvement therefore does not imply a shared direction of host-token change.

The traces illustrate two different paths to a successful result. In the ALFWorld cases described in Section~\ref{sec:amplification}, the agent proceeds from pickup to cleaning and delivery instead of repeatedly putting down the object. The \#2 controller can select held-object actions without calling the host. In the SpreadsheetBench formula case from Section~\ref{sec:inversion}, verification instead keeps the host working: rejection and diagnostic feedback lead to further script revisions before a concrete value is written and accepted. Success here involves additional host work. These examples describe the evaluated workflows, whose rules and control flow also change with delegation. \emph{Improving task performance therefore does not by itself establish greater token efficiency: delegation must be assessed by both what the workflow achieves and how much host inference it requires.}

\section{Conclusion}
\label{sec:conclusion}

In this work, we investigated when high-level decision delegation benefits LLM agents through an empirical study across three workflows and six host models. We organized ten candidate decision functions and evaluated selected single-point and combined substitutions using Jev. Our results show that \textbf{Jev can strengthen an agent, but its value depends on which decisions it takes over and how those decisions shape task execution.} Effective delegation therefore requires choosing which decisions Jev should make for each host and workflow, and assessing the effects on both task performance and host-token use. This motivates more selective, context-aware delegation strategies that account for the host model's capabilities, decision functions, and downstream consequences.

\section*{Limitations}

Our study has several limitations. First, we evaluate three benchmarks, and not all ten candidate decision functions are applicable to every workflow. Future work will extend our evaluation to broader domains, including deep research and software engineering. Second, we examine manually selected, fixed substitutions rather than automatically determining when and where delegation is beneficial. Third, differences in workflow implementations and available context may influence the observed effects despite our additional controls. Finally, our evaluation primarily focuses on end-to-end outcomes, without systematically comparing Jev and host LLMs on decision-level accuracy and robustness under identical states. Further investigation is needed to understand how local decision quality translates into downstream performance.

\bibliography{references}

@article{zhu2026jevagentharness,
  title={JevAgentHarness: Benchmarking, Deploying and Training Decision Models That Share Control with LLMs},
  author={Zhu, Kunlun and Feng, Tao and Han, Xu and Wang, Heng and Yang, Ling and Tang, Xiangru and You, Jiaxuan and Ji, Heng},
  journal={arXiv preprint arXiv:2610.21412},
  year={2026}
}

@article{ling2026faster,
  title={When Faster Decisions Do Not Make Better Agents: An Empirical Study of Jev in Agent Workflows},
  author={Ling, Yuchen and Yu, Shengcheng and Chen, Zhenyu and Fang, Chunrong},
  journal={arXiv preprint arXiv:2610.13092},
  year={2026}
}

@article{vats2026good,
  title={Good Judgment, Bad Advice: De\-ci\-sion-Mod\-el Con\-trol\-lers Raise Agent Cost Without Improving Success},
  author={Vats, Naman},
  journal={arXiv preprint arXiv:\allowbreak{}2610.\allowbreak{}24649},
  year={2026}
}

@article{liu2026actor,
  title={Actor, Judge, or Teacher? Understanding Decision Models in Long-Horizon Agents},
  author={Liu, Haochen and Zhu, Qianhao and Wang, Zixuan and Li, Dingming and Chen, Tongbo and Qiu, Peng and Qu, Yingqi and Lu, Weiming and Xiao, Jun and Shen, Yongliang},
  journal={arXiv preprint arXiv:2611.17167},
  year={2026}
}

@misc{almeida2026jev,
  title={Introducing {System One} Models \& {Jev}},
  author={Almeida, Diogo},
  year={2026},
  month=sep,
  howpublished={TypeSafe AI},
  url={https://typesafe.ai/blog/introducing-system-one-models-and-jev}
}

@article{deusser2026evaluating,
  title={Evaluating and benchmarking the System One model Jev},
  author={Deu{\ss}er, Tobias and Ramamurthy, Rajkumar and Sparrenberg, Lorenz and Sifa, Rafet},
  journal={arXiv preprint arXiv:2609.37647},
  year={2026}
}

@article{kim2023llm,
  title={An llm compiler for parallel function calling},
  author={Kim, Sehoon and Moon, Suhong and Tabrizi, Ryan and Lee, Nicholas and Mahoney, Michael W and Keutzer, Kurt and Gholami, Amir},
  journal={arXiv preprint arXiv:2312.04511},
  year={2023}
}

@article{yao2022react,
  title={React: Synergizing reasoning and acting in language models},
  author={Yao, Shunyu and Zhao, Jeffrey and Yu, Dian and Du, Nan and Shafran, Izhak and Narasimhan, Karthik and Cao, Yuan},
  journal={arXiv preprint arXiv:2210.03629},
  year={2022}
}

@article{shinn2023reflexion,
  title={Reflexion: Language agents with verbal reinforcement learning},
  author={Shinn, Noah and Cassano, Federico and Gopinath, Ashwin and Narasimhan, Karthik and Yao, Shunyu},
  journal={Advances in neural information processing systems},
  volume={36},
  pages={8634--8652},
  year={2023}
}

@article{zhou2023language,
  title={Language agent tree search unifies reasoning acting and planning in language models},
  author={Zhou, Andy and Yan, Kai and Shlapentokh-Rothman, Michal and Wang, Haohan and Wang, Yu-Xiong},
  journal={arXiv preprint arXiv:2310.04406},
  year={2023}
}

@article{madaan2023self,
  title={Self-refine: Iterative refinement with self-feedback},
  author={Madaan, Aman and Tandon, Niket and Gupta, Prakhar and Hallinan, Skyler and Gao, Luyu and Wiegreffe, Sarah and Alon, Uri and Dziri, Nouha and Prabhumoye, Shrimai and Yang, Yiming and others},
  journal={Advances in neural information processing systems},
  volume={36},
  pages={46534--46594},
  year={2023}
}

@inproceedings{zhou2024webarena,
  title={Webarena: A realistic web environment for building autonomous agents},
  author={Zhou, Shuyan and Xu, Frank F and Zhu, Hao and Zhou, Xuhui and Lo, Robert and Sridhar, Abishek and Cheng, Xianyi and Ou, Tianyue and Bisk, Yonatan and Fried, Daniel and others},
  booktitle={International Conference on Learning Representations},
  volume={2024},
  pages={15585--15606},
  year={2024}
}

@article{opsahl2026officeqa,
  title={Officeqa pro: An enterprise benchmark for end-to-end grounded reasoning},
  author={Opsahl-Ong, Krista and Singhvi, Arnav and Collins, Jasmine and Zhou, Ivan and Wang, Cindy and Baheti, Ashutosh and Oertell, Owen and Portes, Jacob and Havens, Sam and Elsen, Erich and others},
  journal={arXiv preprint arXiv:2603.08655},
  year={2026}
}

@article{shridhar2020alfworld,
  title={Alfworld: Aligning text and embodied environments for interactive learning},
  author={Shridhar, Mohit and Yuan, Xingdi and C{\^o}t{\'e}, Marc-Alexandre and Bisk, Yonatan and Trischler, Adam and Hausknecht, Matthew},
  journal={arXiv preprint arXiv:2010.03768},
  year={2020}
}

@article{ma2024spreadsheetbench,
  title={Spreadsheetbench: Towards challenging real world spreadsheet manipulation},
  author={Ma, Zeyao and Zhang, Bohan and Zhang, Jing and Yu, Jifan and Zhang, Xiaokang and Zhang, Xiaohan and Luo, Sijia and Wang, Xi and Tang, Jie},
  journal={Advances in Neural Information Processing Systems},
  volume={37},
  pages={94871--94908},
  year={2024}
}

@inproceedings{kim2026cost,
  title={The cost of dynamic reasoning: Demystifying ai agents and test-time scaling from an ai infrastructure perspective},
  author={Kim, Jiin and Shin, Byeongjun and Chung, Jinha and Rhu, Minsoo},
  booktitle={2026 IEEE International Symposium on High Performance Computer Architecture (HPCA)},
  pages={1--16},
  year={2026},
  organization={IEEE}
}

@article{yang2026skillopt,
  title={Skillopt: Executive strategy for self-evolving agent skills},
  author={Yang, Yifan and Gong, Ziyang and Huang, Weiquan and Yang, Qihao and Zhou, Ziwei and Huang, Zisu and Li, Yan and Gao, Xuemei and Dai, Qi and Liu, Bei and others},
  journal={arXiv preprint arXiv:2605.23904},
  year={2026}
}

@article{li2026jev,
  title={JEV-as-a-judge: Accept when con\-fident, escalate when unsure},
  author={Li, Yubo and Miao, Yidi and Krishnan, Ramayya and Padman, Rema},
  journal={arXiv preprint arXiv:\allowbreak{}2609.\allowbreak{}26550},
  year={2026}
}

@article{zhang2026same,
  title={Same scores, different decisions: Evaluating JEV and language models for legal document understanding},
  author={Zhang, Fan and Chen, Yankai and Xie, Zhuohan and Zhou, Yixi and Peng, Sijia and Fan, Lei and Ji, Xinhua and Zheng, Cunyuan and Shan, Huangyong and Yu, Philip S and others},
  journal={arXiv preprint arXiv:2609.27678},
  year={2026}
}

@article{ibrahim2026evaluating,
  title={Evaluating decision models for text annotation in computational social science},
  author={Ibrahim, Hazem and Zaki, Yasir},
  journal={arXiv preprint arXiv:2609.24574},
  year={2026}
}

@article{jiang2026jev,
  title={Jev-mem: System-one-controlled agentic memory for efficient ai agents},
  author={Jiang, Dongming and Li, Yi and Li, Bingzhe},
  journal={arXiv preprint arXiv:2609.23986},
  year={2026}
}

@article{wu2026reflex,
  title={REFLEX with Jev for efficient selective control in LLM agents},
  author={Wu, Tiantong and Lim, Wei Yang Bryan},
  journal={arXiv preprint arXiv:2609.26532},
  year={2026}
}

@article{liu2026jev,
  title={Is Jev a Good Multilingual Judge?},
  author={Liu, Wei and Fan, Yi and Fu, Xiyan},
  journal={arXiv preprint arXiv:\allowbreak{}2611.\allowbreak{}14861},
  year={2026}
}

@misc{openai2026gpt61sol,
  author = {{OpenAI}},
  title = {Addendum to {GPT-6 Astra} System Card: {GPT-6.1 Sol}},
  year = {2026},
  month = sep,
  howpublished = {OpenAI system card},
  url = {https://cdn.openai.com/pdf/38e3efcf-545e-44cd-99ec-2b7eb395f4cc/oai_GPT_6_1_Sol.pdf}
}

@article{xu2026deepseek,
  title={Deepseek-v4. 1-flash: Pushing the limits of kv cache compression},
  author={Xu, Anyi and Li, B and Lin, Bangcai and Xue, Bing and Xian, BingCheng and Xu, Bingzheng and Wu, Bochao and Zhang, Bowei and Deng, Boyi and Yu, CC and others},
  journal={arXiv preprint arXiv:2609.19969},
  year={2026}
}

@article{qiu2026design,
  title={On the design of qwen3. 8-next architecture: Evaluation, efficiency, and training stability},
  author={Qiu, Zihan and Wang, Zekun and Li, Xiao and Li, Yanpeng and Xu, Yang and Wang, Yixuan and Zhang, Huaqing and Men, Rui and Mao, Bochao and Zhang, Chengruidong and others},
  journal={arXiv preprint arXiv:2608.30320},
  year={2026}
}

@misc{anthropic2026haiku55,
  author = {{Anthropic}},
  title = {System Card: {Claude Haiku 5.5}},
  year = {2026},
  month = oct,
  howpublished = {Anthropic system card},
  url = {https://assets.claude.com/e1080d6bf5ae2018ea3c2f414064be03232f5be5.pdf}
}

@misc{openai2026gpt54mini,
  author = {{OpenAI}},
  title = {Introducing {GPT-5.4 mini and nano}},
  year = {2026},
  month = mar,
  howpublished = {OpenAI},
  url = {https://openai.com/index/introducing-gpt-5-4-mini-and-nano/}
}

@misc{openai2026gpt6solluna,
  author = {{OpenAI}},
  title = {Introducing {GPT-6 Sol and Luna}},
  year = {2026},
  month = sep,
  howpublished = {OpenAI},
  url = {https://openai.com/index/introducing-gpt-6-sol-and-luna/}
}

@article{li2026replacing,
  title={Replacing large language models with jev decision models for low-latency edge service orchestration},
  author={Li, Delong and Wang, Xu and Gong, Haochen and Lang, Rui and Yu, Guangsheng},
  journal={arXiv preprint arXiv:2609.22753},
  year={2026}
}

@article{zhang2026agenticjevnav,
  title={AgenticJevNav: Agentic Object Search and Navigation via Event-Driven Verification},
  author={Zhang, Shuning and Zhang, Shuli and Wang, Yunheng and Lu, Wei and Zhang, Jiaxi and Kang, Yihang and Wang, Tao and Chen, Xiangyu and Feng, Yixiao and Sheng, Kai and others},
  journal={arXiv preprint arXiv:2611.19912},
  year={2026}
}

@article{perifanis2026selmroute,
  title={SeLMRoute: Probabilistic Semantic Evidence for Large Language Model Routing},
  author={Perifanis, Vasilis and Pavlidis, Nikolaos and Symeonidis, Symeon},
  journal={arXiv preprint arXiv:2609.34736},
  year={2026}
}

@article{wang2026jev,
  title={Jev-LiteMAS: Separating Decisions from Generation in Multi-Agent Workflows},
  author={Wang, Kaixiang and Zhou, Zhaijiacheng and Lou, Jiong and Li, Jie},
  journal={arXiv preprint arXiv:2610.21411},
  year={2026}
}

\appendix
% Balance the introductory text in two columns, then place full-width tables below it.
\onecolumn
% Keep appendix table spacing natural instead of stretching to the page bottom.
\raggedbottom
\begin{multicols}{2}
\section{Complete Experimental Results}
\label{sec:appendix}

Tables~\ref{tab:detail-officeqa}--\ref{tab:detail-spreadsheet-tokens} report results for all six hosts under the single-point substitutions and selected combinations studied in this paper. Decision IDs follow Table~\ref{tab:replacement-points}, with their task-specific uses listed in Table~\ref{tab:task-replacements}. The combinations extend selected single-point settings and do not cover all possible subsets of decisions.

\textbf{Reporting conventions.} For each benchmark, task performance and token use appear in separate tables with the same model and configuration order. OfficeQA Pro reports EM and F1, while ALFWorld and SpreadsheetBench report task success. Token tables separate host LLM and Jev usage in thousands of tokens per task. An additional ALFWorld table reports environment steps and ineffective actions per task, including actions that leave the environment unchanged. Each numerical cell contains one metric. Bold marks the highest task scores and lowest host token use, environment steps, and ineffective actions for each host, including ties at the displayed precision. Jev token use is not highlighted.

\textbf{Comparison settings.} For OfficeQA Pro, the protocol-only control retains the evidence ledger without Jev; the verification controls add code recomputation before an LLM or Jev judgment. For ALFWorld, the native host receives two recent steps, while the full-history LLM control and Jev receive the episode history. For SpreadsheetBench, the native workflow has no added decision gate, and LLM verification uses the evidence and revision flow used by Jev verification. These controls answer different questions and should be kept distinct when comparing configurations.

\end{multicols}

\par\addvspace{10pt}
\noindent\begin{minipage}{\textwidth}
  \centering
  \small
  \setlength{\tabcolsep}{2pt}
  \renewcommand{\arraystretch}{1.05}
  \arrayrulecolor{black}
  \begin{tabular*}{\textwidth}{@{\extracolsep{\fill}}lcccccccccccc@{}}
    \toprule
    Configuration & \multicolumn{2}{c}{\shortstack{GPT-5.4\\mini}} & \multicolumn{2}{c}{\shortstack{Qwen3.8-\\Flash}} & \multicolumn{2}{c}{\shortstack{DeepSeek-\\v4.1-flash}} & \multicolumn{2}{c}{\shortstack{GPT-6\\Luna}} & \multicolumn{2}{c}{\shortstack{GPT-6.1\\Sol}} & \multicolumn{2}{c}{\shortstack{Claude\\Haiku 5.5}} \\
    \cmidrule(lr){2-3}\cmidrule(lr){4-5}\cmidrule(lr){6-7}\cmidrule(lr){8-9}\cmidrule(lr){10-11}\cmidrule(lr){12-13}
    & EM & F1 & EM & F1 & EM & F1 & EM & F1 & EM & F1 & EM & F1 \\
    \midrule
    Protocol only & 15.79 & 16.91 & 53.38 & 54.82 & 53.38 & 54.75 & 39.10 & 41.28 & 43.61 & 45.11 & 29.32 & 29.95 \\
    LLM verification & 21.80 & 22.55 & 59.40 & 60.64 & \textbf{54.14} & \textbf{56.38} & 44.36 & 47.23 & 51.13 & 53.02 & 38.35 & 40.22 \\
    \midrule
    Jev \#3 & 16.54 & 16.91 & 54.14 & 54.95 & \textbf{54.14} & 55.50 & 41.35 & 43.79 & 31.58 & 33.77 & 24.81 & 26.71 \\
    Jev \#7 & 18.05 & 19.16 & \textbf{60.15} & \textbf{61.52} & 53.38 & 55.25 & 44.36 & 46.49 & 47.37 & 48.85 & 30.08 & 30.72 \\
    Jev \#8 & 21.80 & 22.92 & 56.39 & 58.14 & \textbf{54.14} & 56.26 & 40.60 & 42.48 & \textbf{51.88} & \textbf{53.88} & 35.34 & 36.71 \\
    \midrule
    Jev \#3+\#8 & 21.05 & 21.54 & 55.64 & 57.33 & 51.13 & 53.00 & 44.36 & 46.12 & 46.62 & 49.50 & \textbf{39.10} & 39.35 \\
    Jev \#7+\#8 & \textbf{22.56} & \textbf{23.93} & 56.39 & 57.63 & 53.38 & 55.25 & \textbf{49.62} & \textbf{51.50} & 46.62 & 48.36 & 32.33 & 33.82 \\
    Jev \#7+\#6+\#8 & 21.80 & 23.67 & 53.38 & 55.00 & 51.88 & 53.37 & 45.11 & 47.12 & 49.62 & 51.25 & \textbf{39.10} & \textbf{40.65} \\
    \bottomrule
  \end{tabular*}
  \captionof{table}{OfficeQA Pro: task performance (\%) on 133 tasks.}
  \label{tab:detail-officeqa}
\end{minipage}\par

\par\addvspace{10pt}
\noindent\begin{minipage}{\textwidth}
  \centering
  \small
  \setlength{\tabcolsep}{2pt}
  \renewcommand{\arraystretch}{1.05}
  \arrayrulecolor{black}
  \begin{tabular*}{\textwidth}{@{\extracolsep{\fill}}lcccccccccccc@{}}
    \toprule
    Configuration & \multicolumn{2}{c}{\shortstack{GPT-5.4\\mini}} & \multicolumn{2}{c}{\shortstack{Qwen3.8-\\Flash}} & \multicolumn{2}{c}{\shortstack{DeepSeek-\\v4.1-flash}} & \multicolumn{2}{c}{\shortstack{GPT-6\\Luna}} & \multicolumn{2}{c}{\shortstack{GPT-6.1\\Sol}} & \multicolumn{2}{c}{\shortstack{Claude\\Haiku 5.5}} \\
    \cmidrule(lr){2-3}\cmidrule(lr){4-5}\cmidrule(lr){6-7}\cmidrule(lr){8-9}\cmidrule(lr){10-11}\cmidrule(lr){12-13}
    & LLM & Jev & LLM & Jev & LLM & Jev & LLM & Jev & LLM & Jev & LLM & Jev \\
    \midrule
    Protocol only & 124.3 & 0.0 & 287.3 & 0.0 & 199.9 & 0.0 & 108.6 & 0.0 & \textbf{70.8} & 0.0 & \textbf{78.2} & 0.0 \\
    LLM verification & 126.1 & 0.0 & 297.4 & 0.0 & 238.0 & 0.0 & 122.5 & 0.0 & 110.1 & 0.0 & 148.5 & 0.0 \\
    \midrule
    Jev \#3 & \textbf{111.3} & 1.3 & \textbf{258.7} & 1.6 & 206.3 & 1.3 & \textbf{107.4} & 1.3 & 122.6 & 1.2 & 80.4 & 1.1 \\
    Jev \#7 & 243.6 & 30.1 & 448.8 & 24.8 & 394.7 & 27.2 & 191.1 & 17.9 & 293.8 & 18.4 & 257.7 & 19.2 \\
    Jev \#8 & 134.2 & 2.7 & 276.7 & 2.5 & 204.7 & 2.3 & 113.4 & 2.6 & 143.3 & 3.4 & 98.7 & 1.7 \\
    \midrule
    Jev \#3+\#8 & 130.9 & 3.6 & 279.9 & 4.2 & \textbf{192.8} & 3.7 & 118.5 & 3.6 & 114.0 & 3.6 & 102.1 & 3.0 \\
    Jev \#7+\#8 & 245.8 & 28.1 & 452.9 & 28.9 & 398.2 & 28.7 & 202.6 & 19.4 & 364.2 & 24.6 & 275.5 & 20.7 \\
    Jev \#7+\#6+\#8 & 196.2 & 33.1 & 421.9 & 30.6 & 417.0 & 36.4 & 225.9 & 22.4 & 198.6 & 20.6 & 285.8 & 26.8 \\
    \bottomrule
  \end{tabular*}
  \captionof{table}{OfficeQA Pro: host LLM and Jev tokens (k per task).}
  \label{tab:detail-officeqa-tokens}
\end{minipage}\par

\par\addvspace{10pt}
\noindent\begin{minipage}{\textwidth}
  \centering
  \small
  \setlength{\tabcolsep}{2pt}
  \renewcommand{\arraystretch}{1.05}
  \arrayrulecolor{black}
  \begin{tabular*}{\textwidth}{@{\extracolsep{\fill}}lcccccc@{}}
    \toprule
    Configuration & \shortstack{GPT-5.4\\mini} & \shortstack{Qwen3.8-\\Flash} & \shortstack{DeepSeek-\\v4.1-flash} & \shortstack{GPT-6\\Luna} & \shortstack{GPT-6.1\\Sol} & \shortstack{Claude\\Haiku 5.5} \\
    \midrule
    Native workflow & 71.64 & 85.82 & 90.30 & 84.33 & 95.52 & 82.09 \\
    Full-history LLM & 79.85 & \textbf{97.76} & \textbf{100.00} & \textbf{98.51} & \textbf{100.00} & \textbf{97.01} \\
    \midrule
    Jev \#2 & 77.61 & 85.82 & 88.06 & 87.31 & 93.28 & 83.58 \\
    Jev \#3 & 73.13 & 87.31 & 88.81 & 85.82 & 97.01 & 83.58 \\
    Jev \#6 & 89.55 & \textbf{97.76} & 99.25 & \textbf{98.51} & \textbf{100.00} & 94.78 \\
    Jev \#7 & 67.16 & 87.31 & 85.07 & 80.60 & 91.79 & 77.61 \\
    Jev \#9 & 72.39 & 88.81 & 86.57 & 85.07 & 94.78 & 83.58 \\
    \midrule
    Jev \#6+\#2 & \textbf{97.01} & 97.01 & \textbf{100.00} & \textbf{98.51} & 99.25 & \textbf{97.01} \\
    Jev \#6+\#7 & 82.09 & 95.52 & 96.27 & 96.27 & 96.27 & 95.52 \\
    Jev \#6+\#2+\#9 & 95.52 & 97.01 & 91.04 & 97.76 & 98.51 & \textbf{97.01} \\
    \bottomrule
  \end{tabular*}
  \captionof{table}{ALFWorld: task performance (\%) on 134 tasks.}
  \label{tab:detail-alfworld}
\end{minipage}\par

\par\addvspace{10pt}
\noindent\begin{minipage}{\textwidth}
  \centering
  \small
  \setlength{\tabcolsep}{2pt}
  \renewcommand{\arraystretch}{1.05}
  \arrayrulecolor{black}
  \begin{tabular*}{\textwidth}{@{\extracolsep{\fill}}lcccccccccccc@{}}
    \toprule
    Configuration & \multicolumn{2}{c}{\shortstack{GPT-5.4\\mini}} & \multicolumn{2}{c}{\shortstack{Qwen3.8-\\Flash}} & \multicolumn{2}{c}{\shortstack{DeepSeek-\\v4.1-flash}} & \multicolumn{2}{c}{\shortstack{GPT-6\\Luna}} & \multicolumn{2}{c}{\shortstack{GPT-6.1\\Sol}} & \multicolumn{2}{c}{\shortstack{Claude\\Haiku 5.5}} \\
    \cmidrule(lr){2-3}\cmidrule(lr){4-5}\cmidrule(lr){6-7}\cmidrule(lr){8-9}\cmidrule(lr){10-11}\cmidrule(lr){12-13}
    & LLM & Jev & LLM & Jev & LLM & Jev & LLM & Jev & LLM & Jev & LLM & Jev \\
    \midrule
    Native workflow & 78.8 & 0.0 & 68.0 & 0.0 & 48.7 & 0.0 & 54.2 & 0.0 & 99.8 & 0.0 & 71.5 & 0.0 \\
    Full-history LLM & 88.7 & 0.0 & 45.8 & 0.0 & 38.1 & 0.0 & 43.8 & 0.0 & 64.8 & 0.0 & 62.2 & 0.0 \\
    \midrule
    Jev \#2 & 51.6 & 12.0 & 49.6 & 10.0 & 32.8 & 9.6 & 32.3 & 12.4 & 62.0 & 11.0 & 47.8 & 14.8 \\
    Jev \#3 & 78.6 & 0.7 & 64.3 & 0.7 & 50.7 & 0.7 & 52.1 & 0.7 & 51.7 & 0.7 & 70.3 & 0.7 \\
    Jev \#6 & 32.2 & 34.1 & 28.7 & 25.2 & 21.5 & 22.5 & 20.8 & 22.8 & 105.2 & 22.6 & 31.4 & 25.6 \\
    Jev \#7 & 103.5 & 10.3 & 73.7 & 2.6 & 57.6 & 2.5 & 61.2 & 3.2 & 150.7 & 3.0 & 88.2 & 3.4 \\
    Jev \#9 & 78.1 & 1.3 & 70.4 & 0.1 & 52.5 & 0.2 & 54.7 & 0.1 & 90.4 & 0.0 & 72.3 & 0.3 \\
    \midrule
    Jev \#6+\#2 & \textbf{12.5} & 36.1 & 16.4 & 37.0 & \textbf{13.6} & 33.5 & \textbf{12.2} & 34.3 & \textbf{46.6} & 34.0 & \textbf{16.7} & 37.7 \\
    Jev \#6+\#7 & 39.9 & 48.5 & 33.5 & 27.8 & 25.8 & 27.1 & 24.5 & 28.4 & 133.8 & 28.2 & 33.5 & 27.8 \\
    Jev \#6+\#2+\#9 & 12.8 & 37.7 & \textbf{16.2} & 37.3 & 27.9 & 22.7 & 12.7 & 35.8 & 47.2 & 35.1 & 18.8 & 37.5 \\
    \bottomrule
  \end{tabular*}
  \captionof{table}{ALFWorld: host LLM and Jev tokens (k per task).}
  \label{tab:detail-alfworld-tokens}
\end{minipage}\par

\par\addvspace{10pt}
\noindent\begin{minipage}{\textwidth}
  \centering
  \small
  \setlength{\tabcolsep}{2pt}
  \renewcommand{\arraystretch}{1.05}
  \arrayrulecolor{black}
  \begin{tabular*}{\textwidth}{@{\extracolsep{\fill}}lcccccccccccc@{}}
    \toprule
    Configuration & \multicolumn{2}{c}{\shortstack{GPT-5.4\\mini}} & \multicolumn{2}{c}{\shortstack{Qwen3.8-\\Flash}} & \multicolumn{2}{c}{\shortstack{DeepSeek-\\v4.1-flash}} & \multicolumn{2}{c}{\shortstack{GPT-6\\Luna}} & \multicolumn{2}{c}{\shortstack{GPT-6.1\\Sol}} & \multicolumn{2}{c}{\shortstack{Claude\\Haiku 5.5}} \\
    \cmidrule(lr){2-3}\cmidrule(lr){4-5}\cmidrule(lr){6-7}\cmidrule(lr){8-9}\cmidrule(lr){10-11}\cmidrule(lr){12-13}
    & Steps & Ineff. & Steps & Ineff. & Steps & Ineff. & Steps & Ineff. & Steps & Ineff. & Steps & Ineff. \\
    \midrule
    Native workflow & 23.54 & 0.45 & 16.10 & 0.05 & 13.51 & 0.09 & 15.92 & 0.03 & 11.87 & 0.01 & 17.72 & 0.10 \\
    Full-history LLM & 22.09 & 2.04 & \textbf{10.59} & 0.07 & \textbf{9.97} & 0.07 & 11.58 & 0.06 & \textbf{9.21} & 0.02 & 13.87 & 0.23 \\
    \midrule
    Jev \#2 & 20.35 & 0.22 & 15.88 & 0.07 & 13.22 & 0.07 & 14.52 & 0.01 & 12.30 & \textbf{0.00} & 17.72 & 0.06 \\
    Jev \#3 & 22.92 & 0.54 & 14.93 & 0.03 & 13.78 & 0.09 & 14.99 & 0.04 & 11.19 & \textbf{0.00} & 17.02 & 0.04 \\
    Jev \#6 & 18.11 & 0.34 & 12.31 & 0.01 & 11.24 & 0.01 & \textbf{11.44} & \textbf{0.00} & 11.13 & \textbf{0.00} & 14.12 & 0.25 \\
    Jev \#7 & 26.48 & 0.78 & 16.57 & 0.04 & 15.22 & 0.15 & 16.75 & 0.04 & 13.09 & \textbf{0.00} & 20.49 & 0.29 \\
    Jev \#9 & 23.72 & 0.51 & 16.07 & 0.04 & 14.47 & 0.10 & 15.95 & 0.02 & 12.34 & \textbf{0.00} & 17.95 & 0.17 \\
    \midrule
    Jev \#6+\#2 & \textbf{13.04} & 0.03 & 12.77 & \textbf{0.00} & 11.46 & \textbf{0.00} & 11.96 & \textbf{0.00} & 11.61 & \textbf{0.00} & 12.95 & \textbf{0.01} \\
    Jev \#6+\#7 & 19.45 & 0.61 & 12.60 & 0.06 & 12.33 & 0.03 & 12.56 & \textbf{0.00} & 12.56 & \textbf{0.00} & \textbf{12.76} & 0.31 \\
    Jev \#6+\#2+\#9 & 13.40 & \textbf{0.01} & 12.63 & \textbf{0.00} & 13.90 & 0.06 & 12.23 & \textbf{0.00} & 11.95 & \textbf{0.00} & 13.06 & 0.02 \\
    \bottomrule
  \end{tabular*}
  \captionof{table}{ALFWorld: environment steps and ineffective actions per task.}
  \label{tab:detail-alfworld-actions}
\end{minipage}\par

\par\addvspace{10pt}
\noindent\begin{minipage}{\textwidth}
  \centering
  \small
  \setlength{\tabcolsep}{2pt}
  \renewcommand{\arraystretch}{1.05}
  \arrayrulecolor{black}
  \begin{tabular*}{\textwidth}{@{\extracolsep{\fill}}lcccccc@{}}
    \toprule
    Configuration & \shortstack{GPT-5.4\\mini} & \shortstack{Qwen3.8-\\Flash} & \shortstack{DeepSeek-\\v4.1-flash} & \shortstack{GPT-6\\Luna} & \shortstack{GPT-6.1\\Sol} & \shortstack{Claude\\Haiku 5.5} \\
    \midrule
    Native workflow & 60.00 & \textbf{89.29} & 84.29 & 86.07 & \textbf{87.86} & 78.93 \\
    LLM verification & 64.64 & 87.14 & 83.57 & 82.14 & 83.93 & 75.36 \\
    \midrule
    Jev \#6 & 58.93 & 88.21 & 85.36 & \textbf{86.79} & 83.57 & \textbf{80.71} \\
    Jev \#7 & 62.50 & 88.57 & \textbf{86.07} & 85.36 & 83.21 & 77.86 \\
    Jev \#8 & \textbf{66.43} & 84.64 & 81.79 & 82.14 & 81.07 & 76.43 \\
    \midrule
    Jev \#7+\#8 & 66.07 & 83.93 & 84.64 & 79.29 & 81.07 & 73.57 \\
    Jev \#6+\#8 & 63.93 & 84.64 & 82.50 & 80.00 & 81.43 & 76.07 \\
    \bottomrule
  \end{tabular*}
  \captionof{table}{SpreadsheetBench: task performance (\%) on 280 tasks.}
  \label{tab:detail-spreadsheet}
\end{minipage}\par

\par\addvspace{10pt}
\noindent\begin{minipage}{\textwidth}
  \centering
  \small
  \setlength{\tabcolsep}{2pt}
  \renewcommand{\arraystretch}{1.05}
  \arrayrulecolor{black}
  \begin{tabular*}{\textwidth}{@{\extracolsep{\fill}}lcccccccccccc@{}}
    \toprule
    Configuration & \multicolumn{2}{c}{\shortstack{GPT-5.4\\mini}} & \multicolumn{2}{c}{\shortstack{Qwen3.8-\\Flash}} & \multicolumn{2}{c}{\shortstack{DeepSeek-\\v4.1-flash}} & \multicolumn{2}{c}{\shortstack{GPT-6\\Luna}} & \multicolumn{2}{c}{\shortstack{GPT-6.1\\Sol}} & \multicolumn{2}{c}{\shortstack{Claude\\Haiku 5.5}} \\
    \cmidrule(lr){2-3}\cmidrule(lr){4-5}\cmidrule(lr){6-7}\cmidrule(lr){8-9}\cmidrule(lr){10-11}\cmidrule(lr){12-13}
    & LLM & Jev & LLM & Jev & LLM & Jev & LLM & Jev & LLM & Jev & LLM & Jev \\
    \midrule
    Native workflow & \textbf{26.4} & 0.0 & 135.8 & 0.0 & 137.2 & 0.0 & \textbf{31.6} & 0.0 & \textbf{46.3} & 0.0 & \textbf{66.0} & 0.0 \\
    LLM verification & 65.5 & 0.0 & \textbf{97.7} & 0.0 & 162.8 & 0.0 & 48.4 & 0.0 & 83.0 & 0.0 & 107.5 & 0.0 \\
    \midrule
    Jev \#6 & 28.9 & 3.1 & 166.5 & 28.1 & 129.5 & 11.6 & 42.3 & 8.6 & 55.6 & 4.5 & 91.2 & 17.4 \\
    Jev \#7 & 35.8 & 5.1 & 105.9 & 5.7 & \textbf{118.2} & 5.8 & 39.0 & 5.8 & 66.1 & 5.1 & 76.9 & 5.7 \\
    Jev \#8 & 51.0 & 5.3 & 154.2 & 4.8 & 201.4 & 3.8 & 52.2 & 5.2 & 72.3 & 4.8 & 139.5 & 7.5 \\
    \midrule
    Jev \#7+\#8 & 60.9 & 9.8 & 126.7 & 10.6 & 182.4 & 10.2 & 61.0 & 11.1 & 83.9 & 9.4 & 137.6 & 12.6 \\
    Jev \#6+\#8 & 54.6 & 11.1 & 167.8 & 31.6 & 172.9 & 21.7 & 74.1 & 17.9 & 87.6 & 11.6 & 168.6 & 33.4 \\
    \bottomrule
  \end{tabular*}
  \captionof{table}{SpreadsheetBench: host LLM and Jev tokens (k per task).}
  \label{tab:detail-spreadsheet-tokens}
\end{minipage}\par

% Continue in two columns directly after the final result table in Appendix A.
\begin{multicols}{2}
\section{Case Studies of Decision Delegation}
\label{sec:case-studies}

We examine four paired execution traces to show how delegated decisions change task execution. The cases illustrate the findings in Sections~\ref{sec:amplification}, \ref{sec:inversion}, \ref{sec:interactions}, and~\ref{sec:resource-use}. Task IDs identify the recorded episodes. The excerpts retain selected actions, outputs, and decision scores, with intervening steps omitted. Jev selects actions or returns acceptance probabilities, while the workflow controller generates diagnostic feedback. These selected cases explain observed behavior rather than measure how often each behavior occurs.

\subsection{Completing a Sequence after Object Pickup}
\label{sec:case-progress}

\textbf{Task and settings.} In ALFWorld task \texttt{test:0053}, GPT-5.4 mini must clean a knife and place it on a countertop. We compare the full-history LLM control with delegation of search (\#6) and held-object progression (\#2). Both receive episode history, while the delegation configuration also assigns the corresponding task guidance to Jev.

\begin{quote}
\small\raggedright
\textbf{Full-history LLM, selected actions:}\\
\texttt{take knife 1 from countertop 2}\\
\texttt{clean knife 1 with sinkbasin 1}\\
Environment: ``Nothing happens.''\\
\texttt{move knife 1 to countertop 2}\\[3pt]
\textbf{Jev \#6+\#2, after host pickup:}\\
\#2: \texttt{go to sinkbasin 1}\\
\#2: \texttt{clean knife 1 with sinkbasin 1}\\
Environment: ``You clean the knife 1 using the sinkbasin 1.''\\
\#2: \texttt{go to countertop 2}\\
\#2: \texttt{move knife 1 to countertop 2}
\end{quote}

The control finds and picks up the knife, but attempts cleaning while still at the countertop. It repeatedly picks up and replaces the object and fails at the 50-step limit. In the delegated run, Jev first selects two countertop search actions, the host picks up the knife, and \#2 then selects navigation, cleaning, return, and placement. The task succeeds in seven environment steps. Host-token use falls from 218,261 to 5,277, with 18,333 Jev tokens in the delegated run. This case illustrates how preserving progress between subgoals can improve success and reduce repeated host inference. The comparison concerns the complete configurations, including their allocation of task guidance.

\subsection{Verification that Prompts a Useful Revision}
\label{sec:case-useful-verification}

\textbf{Task and settings.} In SpreadsheetBench task \texttt{49237}, GPT-5.4 mini must write a middle initial followed by a period in cell F7, leaving it blank when no initial is present. We compare native ReAct with result verification (\#8). The verifier receives the instruction, script, execution output, and saved-workbook evidence. A rejection can keep the host working when it attempts to stop.

\begin{quote}
\small\raggedright
\textbf{First output:} F7 contains an unevaluated formula.\\
\textbf{Jev:} acceptance probability 0.05; reject.\\
\textbf{Controller feedback, excerpt:}\\
``The saved output is not accepted yet, so the task is not finished.''\\
``1 hold a formula with no computed value''\\[3pt]
\textbf{Second output:} F7 is \texttt{None}.\\
\textbf{Jev:} acceptance probability 0.11; reject.\\
\textbf{Third output:} F7 is \texttt{G.}\\
\textbf{Jev:} acceptance probability 0.51; accept.
\end{quote}

The host initially tries to finish with a formula that has no cached value, which the evaluator reads as empty. After the controller refuses this stop attempt and supplies answer-range diagnostics, the host revises the script and eventually writes the required literal value. The delegated run passes, whereas native ReAct fails with an empty evaluated F7. The successful run uses 47,114 host tokens, compared with 25,732 in the native run, plus 4,206 Jev tokens. Thus, verification can improve the outcome by eliciting additional host work rather than by reducing host-token use. The trace shows the combined action of Jev's acceptance gate and the controller's feedback.

\subsection{Verification that Changes a Required Blank}
\label{sec:case-harmful-verification}

\textbf{Task and settings.} In SpreadsheetBench task \texttt{39190}, GPT-6.1 Sol must write \texttt{Delete} in column C when the corresponding value in column A appears in column F, and apply the requested fill color. Nonmatching output cells should remain blank. We again compare native ReAct with result verification (\#8).

\begin{quote}
\small\raggedright
\textbf{First output:} 22 of 35 target cells are blank.\\
\textbf{Jev:} acceptance probability 0.45; reject.\\
\textbf{Controller feedback, excerpt:}\\
``35 target cells; 22 are empty''\\
\texttt{Tabelle1!C6: None}\\[3pt]
\textbf{Revised output:} nonmatching cells contain \texttt{Keep}.\\
\textbf{Jev:} acceptance probability 0.55; accept.\\
\textbf{Final evaluator:}\\
\texttt{value@Tabelle1!C6: gt=None pred='Keep'}
\end{quote}

After rejection and diagnostic feedback, the host fills the nonmatching cells with \texttt{Keep}. Jev accepts the revised workbook, but the task evaluator rejects a cell that should be blank. Native ReAct passes this task. Within the delegated trace, the erroneous value is introduced after verification rejects the earlier output. This does not establish that the entire earlier workbook would have passed, since it was not independently rescored. It does show how a rejection followed by blank-cell diagnostics can prompt an inappropriate revision. Together with the preceding case, this illustrates why acceptance decisions must distinguish missing work from intentional blanks.

\subsection{An Additional Decision that Blocks Progress}
\label{sec:case-interference}

\textbf{Task and settings.} In ALFWorld task \texttt{test:0002}, GPT-5.4 mini must examine an alarmclock under a desklamp. We compare search delegation alone (\#6) with search plus the pickup judgment (\#6+\#7). Both configurations use Jev to select the desk containing the required object.

\begin{quote}
\small\raggedright
\textbf{Jev \#6 alone:}\\
\#6: \texttt{go to desk 1}\\
Host: \texttt{take alarmclock 1 from desk 1}\\
Host: \texttt{use desklamp 1}\\[3pt]
\textbf{Jev \#6+\#7, selected events:}\\
\#6: \texttt{go to desk 1}\\
Host proposes: \texttt{take alarmclock 1 from desk 1}\\
\#7: \texttt{none} (0.54), pickup (0.46).\\
Controller executes: \texttt{look}.
\end{quote}

Search alone completes the task in three steps. With \#7 added, the agent reaches the same desk but repeatedly declines to pick up the visible alarmclock. At one recorded decision, the host explicitly proposes the required pickup, yet Jev assigns higher probability to taking nothing, and the controller replaces the action with \texttt{look}. Across the episode, all 14 pickup judgments select \texttt{none}. Examination and lamp-use actions continue until the 50-step limit, and the task fails. This case exposes a concrete interaction: the added gate blocks the action needed to turn successful search into task completion. Delegating more decisions therefore changes which parts of the host's behavior can reach the environment.

\end{multicols}

\end{document}